\documentclass[10pt,twocolumn,letterpaper]{article}

\usepackage{cvpr}
\usepackage{times}
\usepackage{epsfig}
\usepackage{graphicx}
\usepackage{amsmath}
\usepackage{amssymb}
\usepackage{subfigure}
\usepackage{algorithm}
\usepackage{algorithmic}

\usepackage[pagebackref=true,breaklinks=true,letterpaper=true,colorlinks,bookmarks=false]{hyperref}

\cvprfinalcopy 

\ifcvprfinal\fi
\begin{document}
\title{Deep Representation Learning with Part Loss for Person Re-Identification}
\author{
Hantao Yao, Shiliang Zhang, Yongdong Zhang, Jintao Li, Qi Tian\\
\{yaohantao,zhyd,jtli\}@ict.ac.cn, slzhang.jdl@pku.edu.cn, qi.tian@utsa.edu
}

\maketitle

\begin{abstract}
 Learning discriminative representations for unseen person images is critical for person Re-Identification (ReID). Most of current approaches learn deep representations in classification tasks, which essentially minimize the empirical classification risk on the training set. As shown in our experiments, such representations easily get overfitted on a discriminative human body part among the training set. To gain the discriminative power on unseen person images, we propose a deep representation learning procedure named Part Loss Networks (PL-Net), to minimize both the empirical classification risk and the representation learning risk. The representation learning risk is evaluated by the proposed part loss, which automatically detects human body parts, and computes the person classification loss on each part separately. Compared with traditional global classification loss, simultaneously considering part loss enforces the deep network to learn representations for different parts and gain the discriminative power on unseen persons. Experimental results on three person ReID datasets, \emph{i.e.,} Market1501, CUHK03, VIPeR, show that our representation outperforms existing deep representations.
\end{abstract}

\section{Introduction}\label{sec:intr}

Person Re-Identification (ReID) targets to identify a probe person appeared under multiple cameras. More specifically, person ReID can be regarded as a challenging zero-shot learning problem, because the training and test sets do not share any person in common. Therefore, person ReID requires discriminative representations to depict unseen person images.

\begin{figure}
\begin{center}
\includegraphics[width=0.9\linewidth]{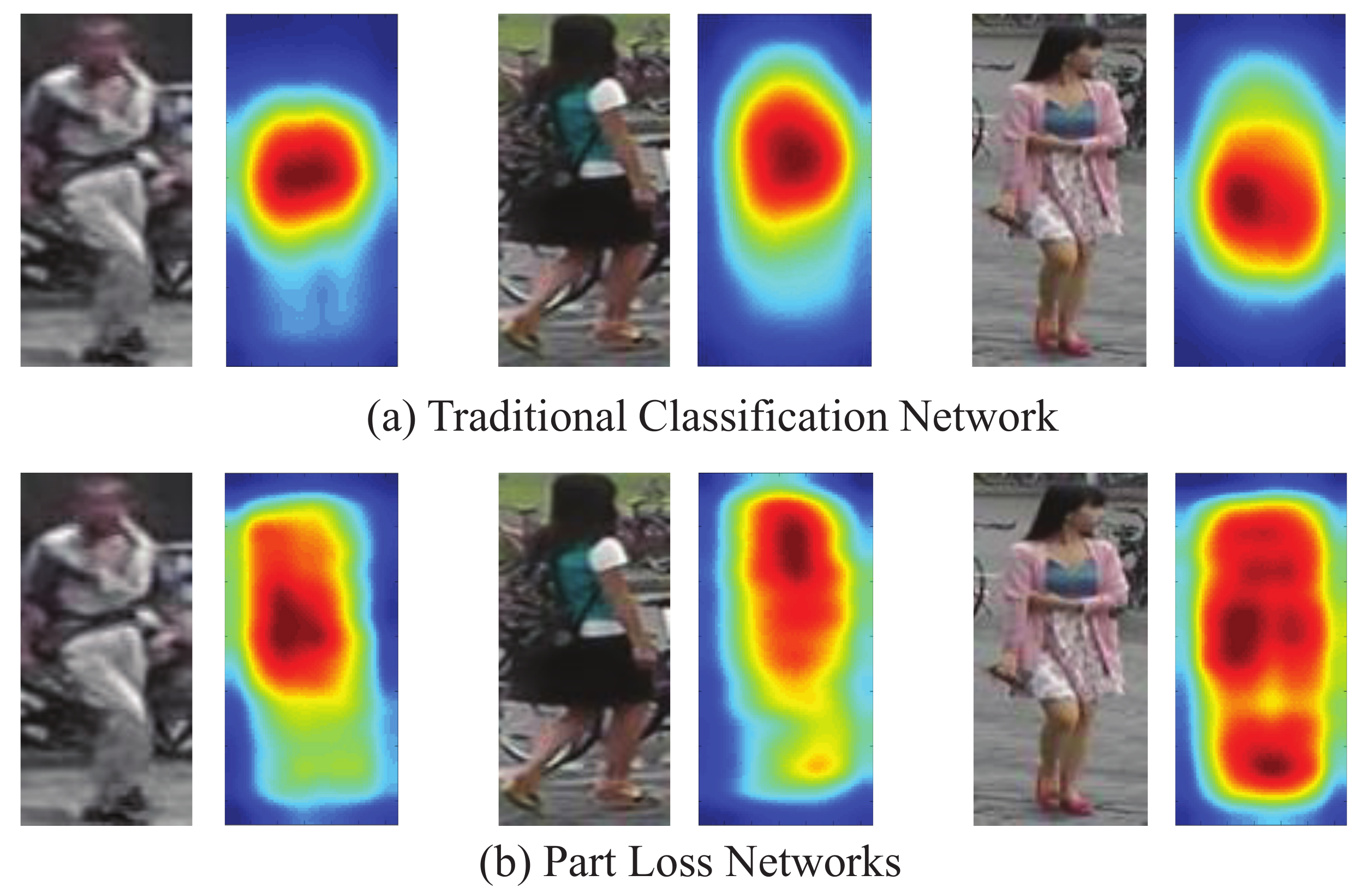}\\
\end{center}
\caption{Saliency maps of CNN learned in traditional classification network (a), and part loss networks (PL-Net) (b). The salient region reveals the body part that the CNN representation focuses on. Representations of our PL-Net are more discriminative to different parts.}
\vspace{-1.7em}
\label{Fig:illustration}
\end{figure}

Existing approaches conquer this challenge by either seeking discriminative metrics~\cite{yu2017cross,liao2015person,zheng2013reidentification,lisanti2015person,ma2013domain,bai2017ssm,pedagadi2013local,xiong2014person,liu2013pop,zheng2013reidentification,chen2015similarity,peng2016unsupervised,zhang2016learning}, or generating discriminative features~\cite{matsukawa2016hierarchical,yang2014salient,farenzena2010person,cheng2011custom,liu2012person,zhao2013unsupervised,zheng2015query,li2014deepreid,zhou2017efficient}. Inspired by the success of Convolutional Neural Network (CNN) in large-scale visual classification~\cite{krizhevsky2012imagenet}, lots of approaches have been proposed to generate representations based on CNN~\cite{chung2017two,li2014deepreid,varior2016gated,xiao2016learning,zhao2014learning,ahmed2015improved,zheng2016discriminatively,su2016deep,zhao2017spindle,wu2016comprehensive,ustinova2015multiregion,zheng2017pose,su2017pose,lin2017consistent,qian2017multi}. For example, several works~\cite{xiao2016learning,zheng2016person,wu2016enhanced} employ deep classification model to learn representations. More detailed reviews on deep learning based person ReID will be given in Sec.~\ref{sec:relatedwork}.

Notwithstanding the success of these approaches, we argue that representations learned by current classification models are not optimal for zero-shot learning problems like person ReID. Most of current deep classification models learn representations by minimizing the classification loss on the training set. This conflicts with the objective of representation learning in person ReID, \emph{i.e.}, gaining high discriminative power to \emph{unseen} person images. Different optimization objectives make current deep representations perform promisingly on classification tasks, but might not be optimal to depict and distinguish unseen person images.

Observations from our experiments are consistent with the above discussions. As shown in Fig.~\ref{Fig:illustration}(a), the representations generated by deep classification model mainly focus on one body region, \emph{i.e.}, the upper body, and ignore the other body parts. This seems reasonable because on the training set, the upper body conveys more distinct clothing cues than the other parts. In order to decrease the classification loss on training data, deep network tends to focus on upper body and ignore the others. However, the other body parts like head, lower-body, and foot are potential to be meaningful for depicting other unseen persons. Ignoring those parts is potential to increases the \emph{risk of representation learning} for unseen data.

The above observations motivate us to study more reliable deep representations for person ReID. We are inspired by the structural risk minimization principle in SVM~\cite{cortes1995support}, which imposes more strict constraint by maximizing the classification margin. Similarly, we enforce the network to learn better representation with extra representation learning risk minimization. Specifically, the representation learning risk is evaluated by the proposed part loss, which automatically generates $K$ parts for an image, and computes the person classification loss on each part separately. In other words, the network is trained to focus on every body part and learn representations for each of them. As illustrated in Fig.~\ref{Fig:illustration}(b), minimizing the person part loss guides the deep network to learn discriminative representations for different body parts. In other words, part loss avoids overfitting on a specific body part, thus decreases the representation learning risk for unseen data.

We propose part loss networks (PL-Net) structure that can be optimized accordingly. As shown in Fig.~\ref{Fig:framework}, part loss networks is composed of a baseline network and an extension to compute the person part loss. It is trained to simultaneously minimize the part loss and the global classification loss. Experiments on three public datasets, \emph{i.e.,} Market1501, CUHK03, VIPeR show PL-Net learns more reliable representations and obtains promising performance in comparison with state-of-the-arts. It also should be noted that, PL-Net is easy to repeat because it only has one important parameter to tune, \emph{i.e.}, the number of generated parts $K$.

Most of previous person ReID works directly train deep classification models to extract image representations. To our best knowledge, this work is an original effort discussing the reasons why such representations are not optimal for person ReID. Representation learning risk and part loss are hence proposed to learn more reliable deep representations to depict unseen person images. The proposed PL-Net is simple but shows promising performance in comparison with the state-of-the-arts. It may also inspire future research on zero-shot learning for person ReID.

\section{Related Work}\label{sec:relatedwork}
The promising performance of CNN on ImageNet classification indicates that classification network extracts discriminative image features. Therefore, several works~\cite{xiao2016learning,zheng2016person,wu2016enhanced} fine-tuned the classification networks on target datasets as feature extractors for person ReID. For example, Xiao~\etal~\cite{xiao2016learning} propose a novel dropout strategy to train a classification model with multiple datasets jointly. Wu~\etal~\cite{wu2016enhanced} combine the hand-crafted histogram features and deep features to fine-tune the classification network.

Besides of classification network, siamese network and triplet network are two popular networks for person ReID. The siamese network takes a pair of images as input, and is trained to verify the similarity between those two images~\cite{zheng2016discriminatively,wu2016personnet,ahmed2015improved,varior2016siamese,yi2014deep,shi2016embedding}. Ahmed~\etal\cite{ahmed2015improved} and Zheng~\etal~\cite{zheng2016discriminatively} employ the siamese network to infer the description and a corresponding similarity metric simultaneously. Shi~\etal\cite{shi2016embedding} replace the Euclidean distance with \emph{Mahalanobis} distance in the siamese network. Varior~\etal~\cite{varior2016siamese} combine the LSTM and siamese network for person ReID. Some other works~\cite{su2016deep,cheng2016person,liu2016end} employ triplet networks to learn the representation for person ReID. Cheng~\emph{et al.}~\cite{cheng2016person} propose a multi-channel parts-based CNN model for person ReID. Liu~\emph{et al.}~\cite{liu2016end} propose an end-to-end Comparative Attention Network to generate image description. Su~\emph{et al.}~\cite{su2016deep} propose a semi-supervised network trained by triplet loss to learn human semantic attributes.

Recently, many works generate deep representation from local parts~\cite{su2017pose,zhao2017spindle,li2017learning,zhao2017deeply}. For example, Su~\etal~\cite{su2017pose}, and Zhao~\etal~\cite{zhao2017spindle} firstly extract human body parts with fourteen body joints, then fuse the features extracted on body parts. Different from \cite{su2017pose} and \cite{zhao2017spindle}, Li~\etal~\cite{li2017learning} employ Spatial Transform Network (STN)~\cite{jaderberg2015spatial} for part localization, and propose Multli-Scale Context-Aware Network to infer representations on the generated local parts.

By analyzing the difference between image classification and person ReID, we find that the representations learned by existing deep classification models are not optimal for person ReID. Therefore, we consider extra representation learning risk and person part loss for deep representation learning. Our work also considers local parts cues for representation learning. Different from existing algorithms~\cite{su2017pose,zhao2017spindle,li2017learning}, part loss networks (PL-Net) automatically detects human parts and does not need extra annotation or detectors, thus is more efficient and easier to implement.

\section{Methodology}\label{sec:methodology}

\subsection{Formulation} \label{sec: formulation}
Given a probe person image $I_{q}$, person ReID targets to return the images containing the identical person in $I_{q}$ from a gallery set $G$. We denote the gallery set as $G=\{I_{i}\}, i\in[1,m]$, where $m$ is the total number of person images. Person ReID can be tackled by learning a discriminative feature representation $\textbf{f}$ for each person image from a training set $T$. Therefore, the probe image can be identified by matching its $\textbf{f}_q$ against the gallery images.

Suppose the training set contains $n$ labeled images from $C$ persons, we denote the training set as $T=\{I_{i}, {y}_{i}\} , i\in[1,n], {y}_{i}\in [1, C]$, where $I_{i}$ is the $i$-th image and ${y}_{i}$ is its person ID label. Note that, person ReID assumes the training and gallery sets contain distinct persons. Therefore, person ReID can be regarded as a zero-shot learning problem, \emph{i.e.}, the ID of probe person is not included in the training set.

Currently, some methods~\cite{xiao2016learning,zheng2016discriminatively,wu2016enhanced} fine-tune a classification-based CNN to generate the feature representation. The feature representation learning can be formulated as updating the CNN network parameter $\theta$ by minimizing the empirical classification risk of representation $\textbf{f}$ on $T$ through back prorogation. We denote the empirical classification risk on $T$ as,
\begin{equation}
\mathcal{J}=\frac{1}{n}[\sum_{i=1}^{n} L^{g}(\hat{y}_i)],
\label{eq:ori_softmaxloss}
\end{equation}
where $\hat{y}_i$ is the predicted classification score for the $i$-th training sample, and $L^{g}(\cdot)$ computes the classification loss for each training image. We use the superscript $^g$ to denote it is computed on the global image. The predicted classification score $\hat{y}_i$ can be formulated as, \emph{i.e.},
\begin{equation}\label{eq:classpredict}
\hat{y}_i=\mathbf{W}\textbf{f}_i+b,
\end{equation}
where $\mathbf{W}$ denotes the parameter of the classifier in CNN, \emph{e.g.}, the weighting matrix in the fully connected layer.

Given a new image $I_{q}$, its representation $\textbf{f}_{q}$ is hence extracted by CNN with the updated parameter $\theta$, \emph{i.e.},
\begin{equation}
\textbf{f}_{q}=\textbf{CNN}_{\theta}(I_{q}).
\end{equation}

It can be inferred from Eq.~\eqref{eq:ori_softmaxloss} and Eq.~\eqref{eq:classpredict} that, to improve the discriminative power of $\textbf{f}_{i}$ during training, a possible way is to restrict the classification ability of $\mathbf{W}$. In another word, a weaker $\mathbf{W}$ would enforce the network to learn more discriminative $\textbf{f}_{i}$ to minimize the classification error. This motivates us to introduce a baseline CNN network with weaker classifiers. Details of this network will be given in Sec.~\ref{sec:basenet}

It also can be inferred from Eq.~\eqref{eq:ori_softmaxloss} that, minimizing the empirical classification risk on $T$ results in a discriminative representation $\textbf{f}$ for classifying the seen categories in $T$. For example in Fig.~\ref{Fig:illustration}(a), the learned representations focus on discriminative parts for training set. However, such representations lack the ability to describe other parts like head, lower-body, and foot which could be meaningful to distinguish an unseen person. Therefore, more parts should be depicted by the network to minimize the risk of representation learning for unseen data.

Therefore, we propose to consider the representation learning risk, which tends to make the CNN network learn discriminative representation for each part of the human body. We denote the representation of each body part as $\textbf{f}^{k}, k\in [1, K]$, where $K$ is the total number of parts. The representation learning risk $\mathcal{P}$ can be formulated as,
\begin{equation}\label{eq:reprisk}
\mathcal{P}=\frac{1}{K}\sum_{k=1}^{K}\frac{1}{n}[\sum_{i=1}^{n}L^{p}(\hat{y}_{i}^k)],
\end{equation}
where $L^{p}(\cdot)$ computes the part loss, \emph{i.e.}, the classification loss on each part. $\hat{y}_{i}^k$ is the predicted person classification score for the $i$-th training sample by the representation of $k$-th part. $\hat{y}_{i}^k$ is computed with,
\begin{equation}
\hat{y}_i^{k}=\textbf{W}^k\textbf{f}_{i}^{k}+b^{k},
\end{equation}
where $\mathbf{W}^k$ denotes the classifier for the representation of the $k$-th part.

The representation learning risk monitors the network and enforces it to learn discriminative representation for each part. It shares a certain similarity with the structural risk minimization principle in SVM~\cite{cortes1995support}, which also imposes more strict constraints to enforce the classifier to learn better discriminative power.

The final part loss networks (PL-Net) model could be inferred by minimizing the empirical classification risk and the representation learning risk simultaneously, \emph{i.e.},
\begin{equation}
\theta = \arg\min(\mathcal{J}+\mathcal{P}).
\label{eq:final_loss}
\end{equation}

In the following parts, we proceed to introduce the part loss networks and the computation of part loss.

\subsection{Part Loss Networks} \label{sec:basenet}

\begin{figure}
\begin{center}
\includegraphics[width=0.9\linewidth]{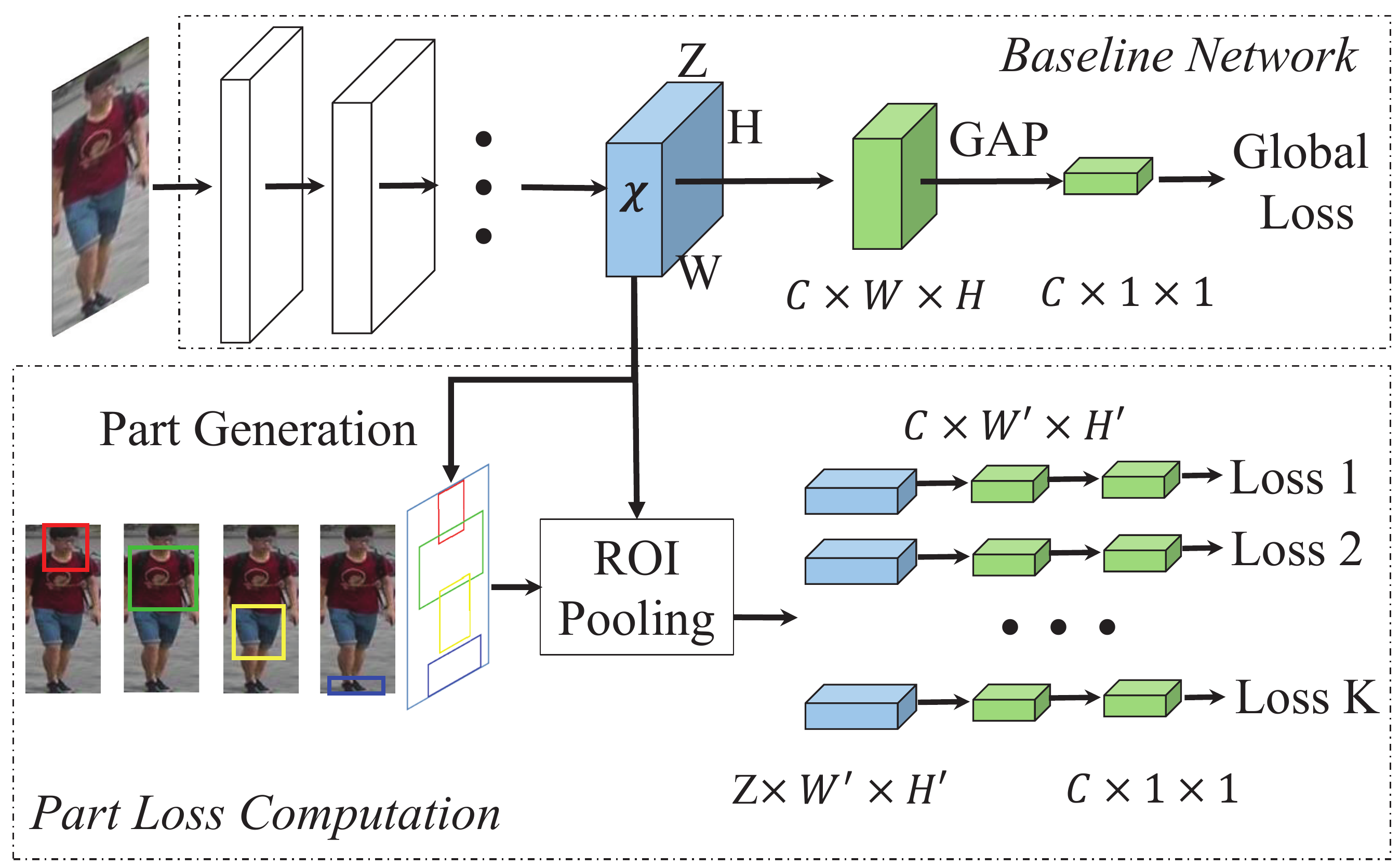}
\end{center}
\caption{Overview of part loss networks (PL-Net), which is composed of a baseline network and a part loss computation extension. ``GAP" denotes the Global Average Pooling. Given an input image, we firstly extract its feature maps $\mathcal{X}$, then compute the global loss and person part loss based on $\mathcal{X}$. The person part loss is computed on $K$ parts generated with an unsupervised method.}
\vspace{-1em}
\label{Fig:framework}
\end{figure}

Most of the deep learning-based person ReID methods treat the Alexnet~\cite{krizhevsky2012imagenet}, GoogLeNet~\cite{szegedy2015going}, and Residual-50~\cite{he2015deep} as the baseline network. Given an image, these networks firstly use several convolutional layers to generate the feature representation, then infer fully-connected layers for classification. Therefore, these networks essentially consist of feature extraction and classification modules.

As discussed in Sec.~\ref{sec: formulation}, weaker classifiers should be used to improve the discriminative power of the learned representations. Moreover, the massive parameters in fully-connected layers may make the network prone to overfitting, especially for small-scale person ReID training sets.

We thus propose a simpler classifier in our baseline network. Our baseline network replaces the fully-connected layers with a convolutional layer and a Global Average Pooling (GAP) layer~\cite{lin2013network}. As shown in Fig.~\ref{Fig:framework}, the convolutional layer directly generates $C$ activation maps explicitly corresponding to $C$ classes. Then GAP hence generates the classification score for each class, \emph{i.e.},
\begin{equation}
s_{c} = \frac{1}{W\times H}\sum_{h=1}^{H}\sum_{w=1}^{W} \mathcal{C}_{c}(h,w),
\label{Eq:gap}
\end{equation}
where $s_{c}$ is the average response of the $c$-th activation map $\mathcal{C}_{c}$ with size $W\times H$, and  $\mathcal{C}_{c}(h,w)$ denotes the activation value on the location $(h,w)$ on $\mathcal{C}_{c}$. $s_{c}$ is hence regarded as the classification score for the $c$-th class. As GAP contains no parameter to learn, it avoids over-fitting and makes the network more compact. We replace FC with GAP because GAP has weak discriminative power and thus needs a powerful feature to ensure the classification accuracy. This encourages the end-to-end training to better focus on feature learning.

According to Eq.~\eqref{eq:final_loss}, our representation is learned to minimize both the empirical classification risk and the representation learning risk. The empirical classification risk is evaluated by the classification loss on the training set. The representation learning risk is evaluated by counting the classification loss on each body part. We thus extend the baseline network accordingly to make it can be optimized by these two types of supervisions. The overall network is shown in Fig.~\ref{Fig:framework}. During training, it computes a person part loss and a global loss with two branches.

Specifically, part loss networks (PL-Net) processes the input image and generates feature maps. We denote the feature maps of the last convolutional layer before the classification module as $\mathcal{X}\in \mathcal{R}^{Z\times H\times W}$. For example, $Z$=1024, $H$=16, $W$=8 when we input $512\times256$ sized image into the baseline network modified from GoogLeNet~\cite{szegedy2015going}. After obtaining $\mathcal{X}$, the global loss is calculated as,
\begin{equation}
L^{g}(\hat{y}_i)=-\sum_{c=1}^{C}1\{y_{i}=c\}\log\frac{e^{\hat{y}_{i}}}{\sum_{l=1}^{C}e^{\hat{y}_{l}}}.
\label{Eq:globalloss}
\end{equation}

The part loss is computed on each automatically generated part to minimize the representation learning risk. The network first generates $K$ person parts based on $\mathcal{X}$ in an unsupervised way. Then part loss is computed on each part by counting its classification loss. The following part gives details of the unsupervised part generation and part loss computation.

\subsection{Person Part Loss Computation}

Person parts can be extracted by various methods. For instance, detection models could be trained with part annotations to detect and extract part locations. However, those methods~\cite{zhao2017spindle,zheng2017pose} require extra annotations that are hard to collect. We thus propose an unsupervised part generation algorithm that can be optimized together with the representation learning procedure.

\begin{figure}
\begin{center}
\includegraphics[width=1\linewidth]{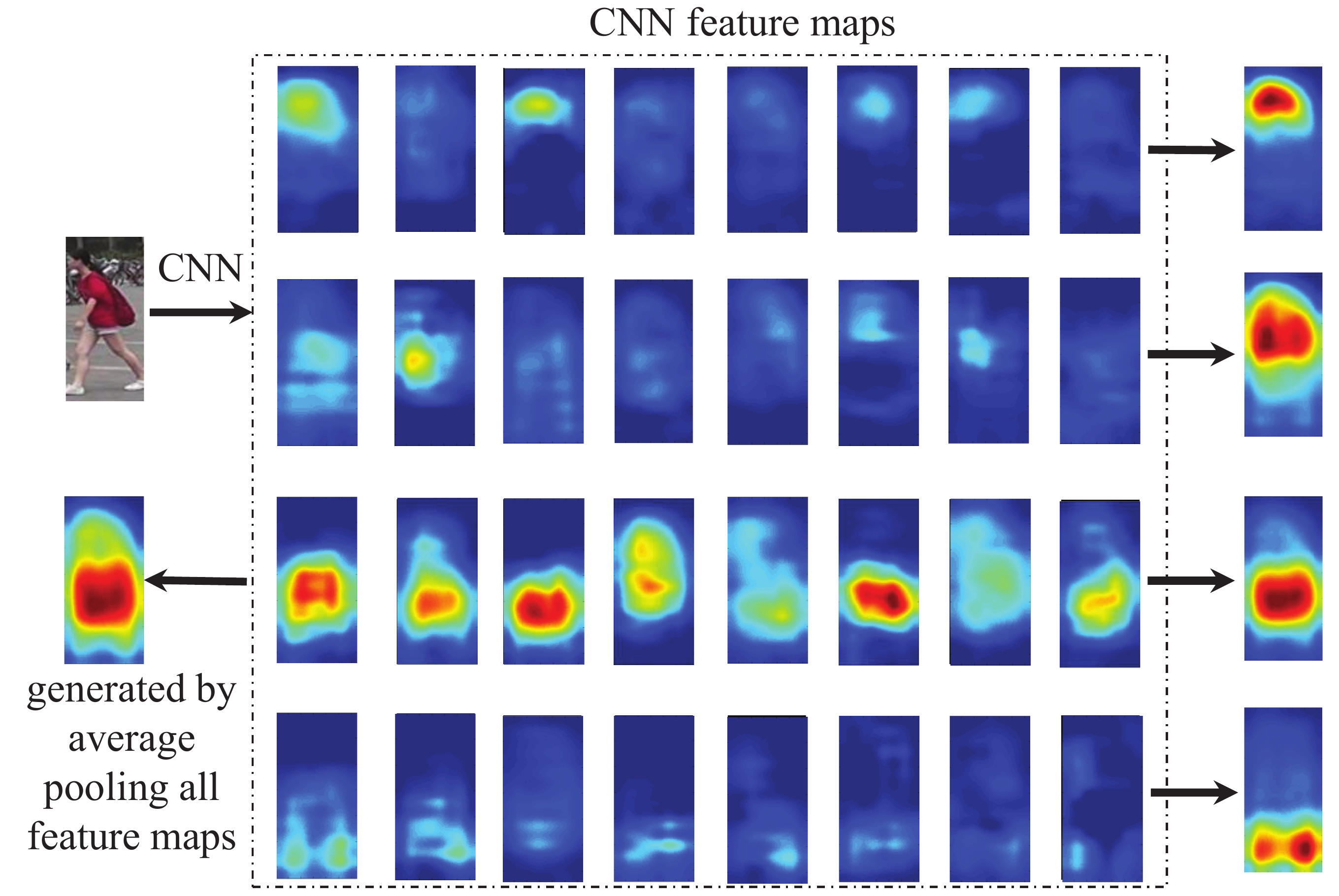}
\end{center}
\caption{Examples of CNN feature maps and generated saliency maps. The saliency map generated on all feature maps focuses on one part and suppresses the activations on other parts. The four saliency maps on the right side are generated by average pooling four clusters of feature maps, respectively. They clearly indicate different part locations.}
\vspace{-1em}
\label{Fig:motivation_personpart}
\end{figure}

Previous work~\cite{wei2016selective} shows that simply average pooling the feature maps of convolutional layers generates a saliency map. The saliency essentially denotes the ``focused'' regions by the neural network. Fig.~\ref{Fig:motivation_personpart} shows several feature maps generated by a CNN trained in the classification task. It can be observed that, the lower part of the body has substantially stronger activations. There exist some feature maps responding to the other parts like head and upper body, but their responses are substantially weaker. As illustrated in Fig.~\ref{Fig:motivation_personpart}, simply average pooling all feature maps shows the discriminative region and suppresses the other regions.

Although the responses on different parts are seriously imbalanced, they still provide cues of different part locations. By clustering feature maps based on the locations of their maximum responses, we can collect feature maps depicting different body parts. Individually average pooling those feature map clusters indicates the part locations. As shown in Fig.~\ref{Fig:motivation_personpart}, the four saliency maps on the right side focus on head, upper body, lower body, and foot, respectively. This might be because the appearances of head, lower body, and foot differs among training persons, thus CNN still learns certain neurons to depict them.

The above observation motivates our unsupervised part generation. Assume that we have got the feature map $\mathcal{X}$, we first compute the position of maximum activation on each feature map, denoted as $(h_{z},w_{z}), z\in[1,Z]$,
\begin{equation}
(h_{z},w_{z})=\arg\max_{h,w} \mathcal{X}_{z}(h,w),
\end{equation}
where $\mathcal{X}_{z}(h,w)$ is the activation value on location $(h,w)$ in the $z$-th channel of $\mathcal{X}$. We then cluster those locations $(h,w)$ into $K$ groups using L2 distance. As the images in current person ReID datasets are cropped and coarsely aligned, we could simply perform clustering only according to the vertical location $h$.

\begin{figure}
\begin{center}
\includegraphics[width=1\linewidth]{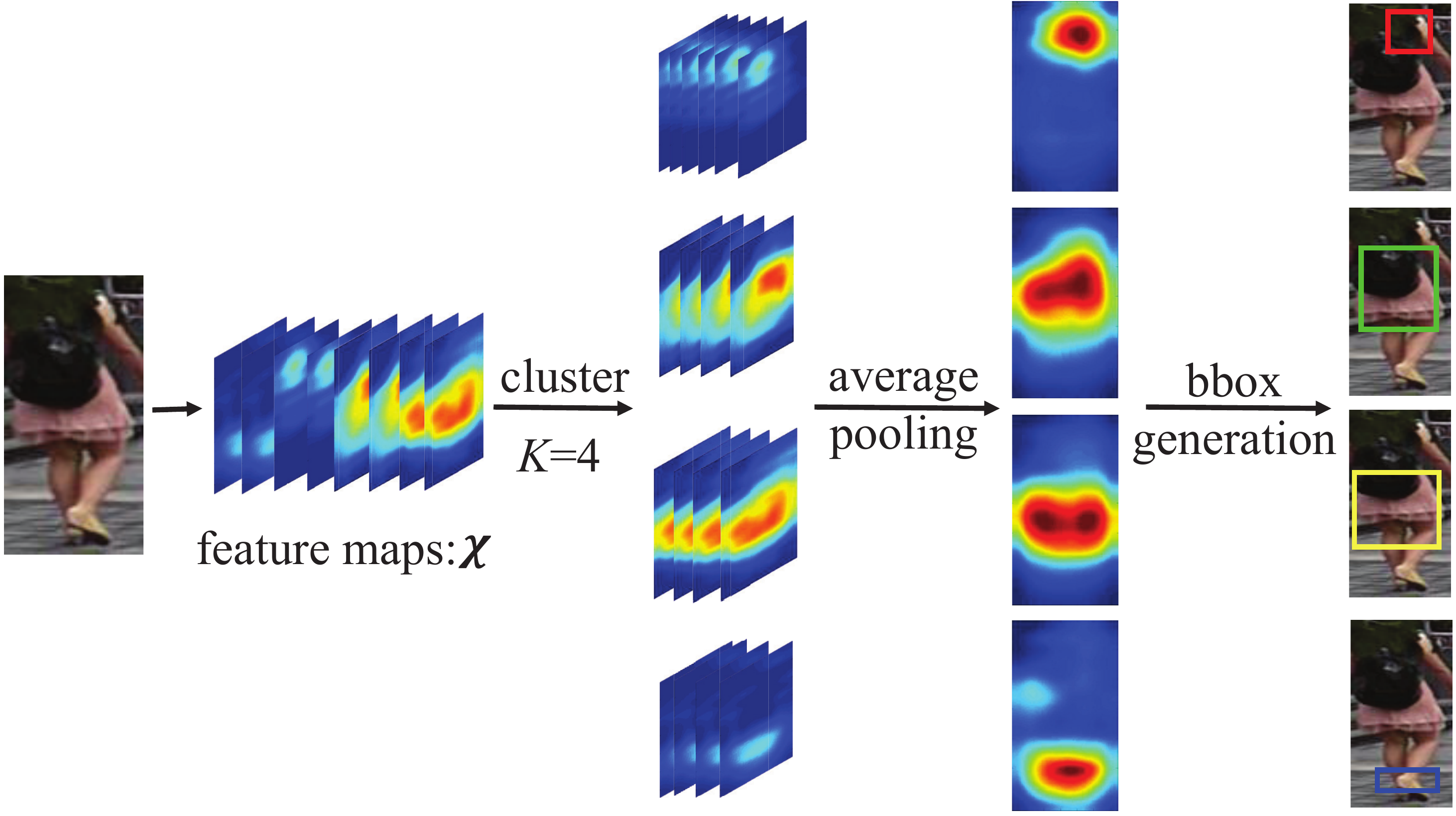}
\end{center}
\caption{Illustration of the procedure for unsupervised person part generation.}
\vspace{-1em}
\label{Fig:personpart}
\end{figure}

After grouping all feature maps into $K$ clusters, we generate one part bounding box from each cluster. Specifically, we average pooling the feature maps in each cluster and apply the max-min normalization to produce the saliency map. A threshold, \emph{e.g.}, 0.5, is set to turn each saliency map into a binary image. In other words, we consider a pixel as foreground if its value is larger than the threshold. For each binary image, we treat its minimum enclosing rectangle as the part bounding box. This procedure is illustrated in Fig.~\ref{Fig:personpart}.

After obtaining the part bounding boxes, we proceed to compute the part loss. Inspired by Fast R-CNN~\cite{girshick2015fast}, we employ the RoI pooling to convert the responses of $\mathcal{X}$ inside each part bounding box into a new feature map $\mathcal{X}^{k}\in \mathcal{R}^{Z\times H^{'}\times W^{'}}$ with a fixed spatial size, \emph{e.g.,} $H^{'} = W^{'} = 4$ in this work. Based on those feature maps, we compute the part loss $L^{p}(\cdot)$ for $k$-th part with a similar procedure of global loss computation, \emph{i.e.},
\begin{equation}
L^{p}(\hat{y}_{l}^k)=-\sum_{c=1}^{C}1\{y_{i}=c\}\log\frac{e^{\hat{y}_{i}^k}}{\sum_{l=1}^{C}e^{\hat{y}_{l}^k}}.
\label{Eq:partkloss}
\end{equation}
Similar to the notations in Eq.~\eqref{eq:reprisk}, $\hat{y}_{i}^k$ is the predicted person classification score of the $i$-th training sample based on the representation of its $k$-th part.

The generated parts are updated on each iteration of network training. It should be noted that, the accuracy of our unsupervised part generation is related with the representation learning performance. For example in Fig.~\ref{Fig:motivation_personpart}, if more neurons are trained to depict parts like head and foot during representation learning, more feature maps would focus on these parts. This in turn improves the feature maps clustering and results in more accurate bounding boxes for these parts. During this procedure, the part generation and representation learning can be jointly optimized.

\begin{figure}
\begin{center}
\includegraphics[width=0.89\linewidth]{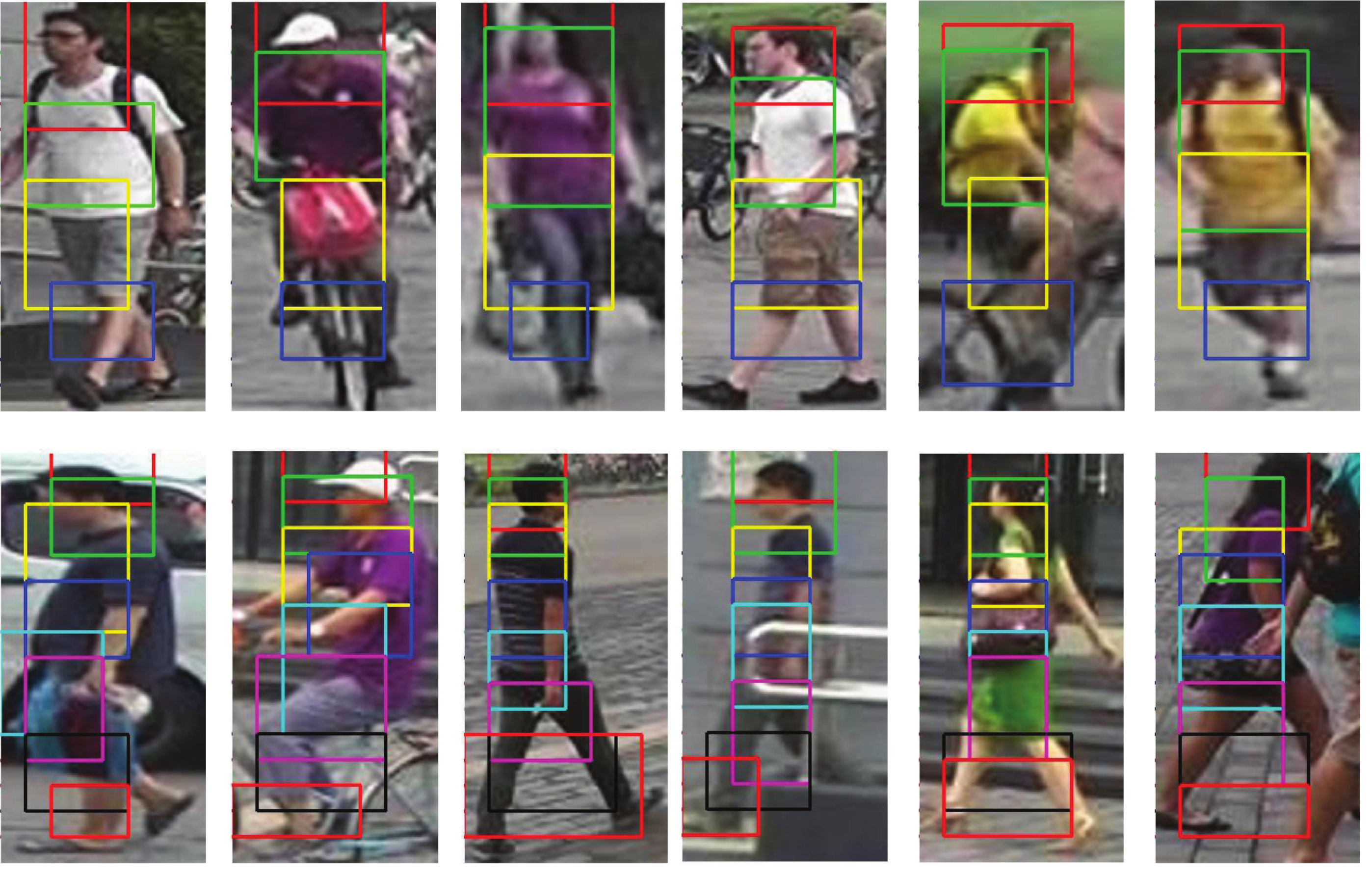}
\end{center}
\caption{Samples of generated part bounding boxes. The first and second row correspond to $K=4$ and $K=8$, respectively.}
\label{Fig:person_part_sample}
\vspace{-1em}
\end{figure}

Examples of generated parts are shown in Fig.~\ref{Fig:person_part_sample}. As shown in Fig.~\ref{Fig:person_part_sample}, the bounding boxes cover important body parts. For the case with $K$=4, the generated four parts coarsely cover the head, upper body, lower body, and legs, respectively. For the case that $K$=8, most of generated parts distribute on the human and cover more detailed parts.

\subsection{Person ReID}

On the testing phase, we extract feature representation from the trained part loss networks for person ReID. We use the feature maps $\mathcal{X}$ to generate the global and part representations for similarity computation.

Given a person image $I$, we firstly resize it to the size of $512\times256$, then input it into the network to obtain feature maps $\mathcal{X}$. We hence compute the global representation $\textbf{f}^{(g)}$ with Eq.~\eqref{Eq:global_descriptor},
\begin{equation}
\textbf{f}^{(g)}=[f_{1},...,f_{z},...f_{Z}],
\label{Eq:global_descriptor}
\end{equation}
\begin{equation}
f_{z}=\frac{1}{W\times H}\sum_{h=1}^{H}\sum_{w=1}^{W} \mathcal{X}_{z}(h,w).
\end{equation}

For the part representation, we obtain the feature maps after RoI pooling for each part, denoted as $\mathcal{X}^{k}\in \mathcal{R}^{Z\times 4\times 4}, k\in[1,K]$. For each $\mathcal{X}^{k}$, we calculate the part description $\textbf{f}^{k}$ in similar way with Eq.~\eqref{Eq:global_descriptor}. The final representation is the concatenation of global and part representations, \emph{i.e.},
\begin{equation}
\textbf{f}=[\textbf{f}^{(g)},\textbf{f}^{1},...,\textbf{f}^{K}].
\end{equation}

\section{Experiments}\label{sec:exp}

\subsection{Datasets}
We verify the proposed part loss networks (PL-Net) on three datasets: VIPeR~\cite{gray2007evaluating}, CUHK03~\cite{li2014deepreid}, and Market1501~\cite{zheng2015scalable}. VIPeR~\cite{gray2007evaluating} contains 632 identities appeared under two cameras. For each identity, there is one image for each camera. The dataset is split randomly into equal halves and cross camera search is performed to evaluate the algorithms.

CUHK03~\cite{li2014deepreid} consists of  14,097 cropped images from 1,467 identities. For each identity, images are captured from two cameras and there are about 5 images for each view. Two ways are used to produce the cropped images, \emph{i.e.}, human annotation and detection by Deformable Part Model (DPM)~\cite{felzenszwalb2010object}. Our evaluation is based on the human annotated images. We use the standard experimental setting~\cite{li2014deepreid} to select 1,367 identities for training, and the rest 100 for testing.

Market1501~\cite{zheng2015scalable} contains 32,668 images from 1,501 identities, and each image is annotated with a bounding box detected by DPM. Each identity is captured by at most six cameras. We use the standard training, testing, and query split provided by the authors in~\cite{zheng2015scalable}. The Rank-1, Rank-5, Rank-10 accuracies are evaluated for VIPeR and CUHK03. For Market1501, we report the Rank-1 accuracy and mean Average Precision (mAP).

\subsection{Implementation Details}
We use Caffe~\cite{jia2014caffe} to implement and train the part loss networks (PL-Net). The baseline network is modified from second version of GoogLeNet~\cite{ioffe2015batch}. Following the \emph{inception5b/output} layer, an $1\times 1$ convolutional layer with the output of $\mathcal{C}$ channels is used  to generate the category confidence map. For the training, we use the pre-trained model introduced in~\cite{lim0606} to initialize the PL-Net, and use a step strategy with mini-batch Stochastic Gradient Descent (SGD) to train the neural networks on Tesla K80 GPU. Parameters like the maximum number of iterations, learning rate, step size, and gamma are set as 50,000, 0.001, 2500, and 0.75, respectively. For the person images, we first resize their size to $512\times 256$, and then feed their into the PL-Net for training and testing.

\begin{figure}
\begin{center}
\includegraphics[width=1\linewidth]{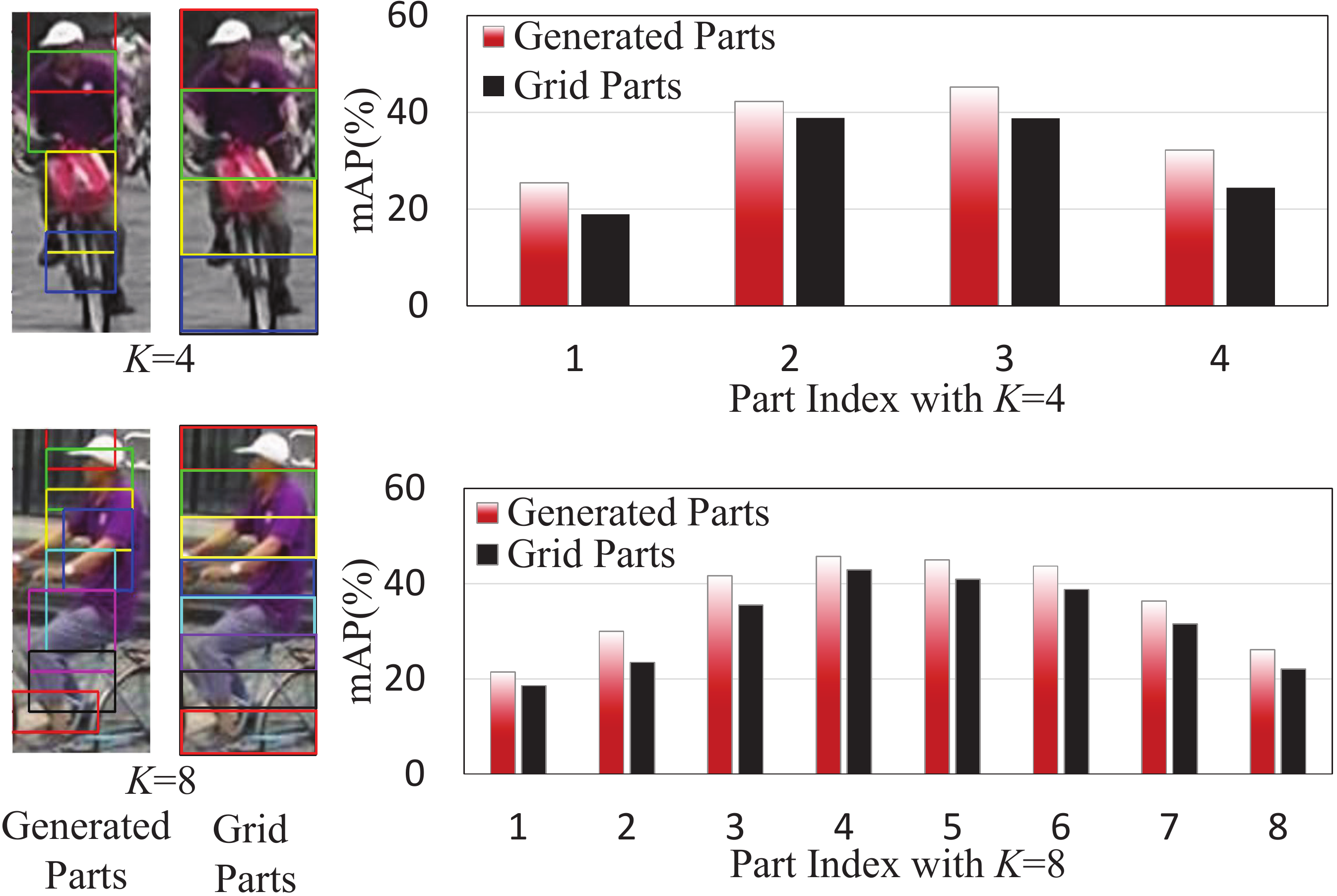}
\end{center}
\caption{Performance comparison of representations learned on generated parts and fixed grid parts on Market1501.}
\vspace{-1em}
\label{Fig:benefits_local_description}
\end{figure}

\subsection{Performance of Learned Representations}

\emph{Accuracy of Part Generation:} One key component of our representation learning is the person part generation. As existing person ReID datasets do not provide part annotations, it is hard to quantify the results. To demonstrate that our generated parts are reasonable, we compare the representations learned by CNN trained with part loss using the \emph{generated parts} and \emph{fixed grid parts}, respectively. As shown on the left side of Fig.~\ref{Fig:benefits_local_description}, we generate grid parts by equally dividing an image into horizontal stripes following previous works~\cite{liao2015person,xiong2014person}. In Fig.~\ref{Fig:benefits_local_description}, the generated parts get substantially higher accuracy than the fixed grid parts for $K$ = 4 and 8, respectively. This conclusion is reasonable, because the generated parts cover most of the human body and filter the clustered backgrounds. It also can be observed that, part representations extracted from the center parts of human body, \emph{e.g.}, parts with index =4 and 5 for $K$=8, get higher accuracies. This might be because the center of human body generally presents more distinct clothing cues. Table ~\ref{Tab:comparison_grid} compares the final global-part representations learned with fixed grid parts and our generated parts. It is clear that, our generated parts perform substantially better.

\begin{figure}
\begin{center}
\includegraphics[width=0.9\linewidth]{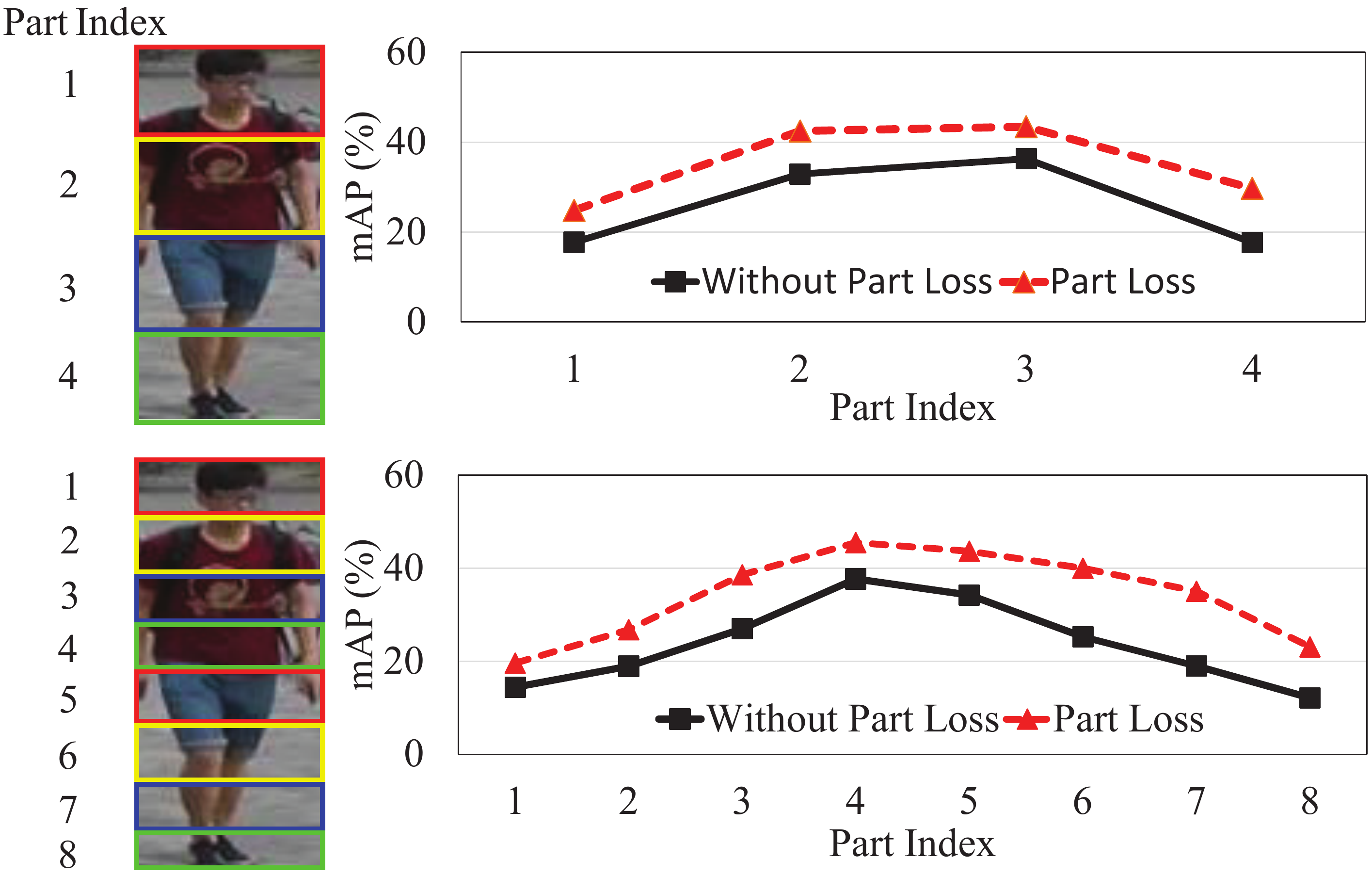}
\end{center}
\caption{Performance of part representations learned with and without part loss on Market1501. We use fixed grid parts in this experiment with $K$=4 and 8, respectively.}
\label{Fig:local_description}
\end{figure}

\begin{figure}
 \centering
\includegraphics[width=0.9\linewidth]{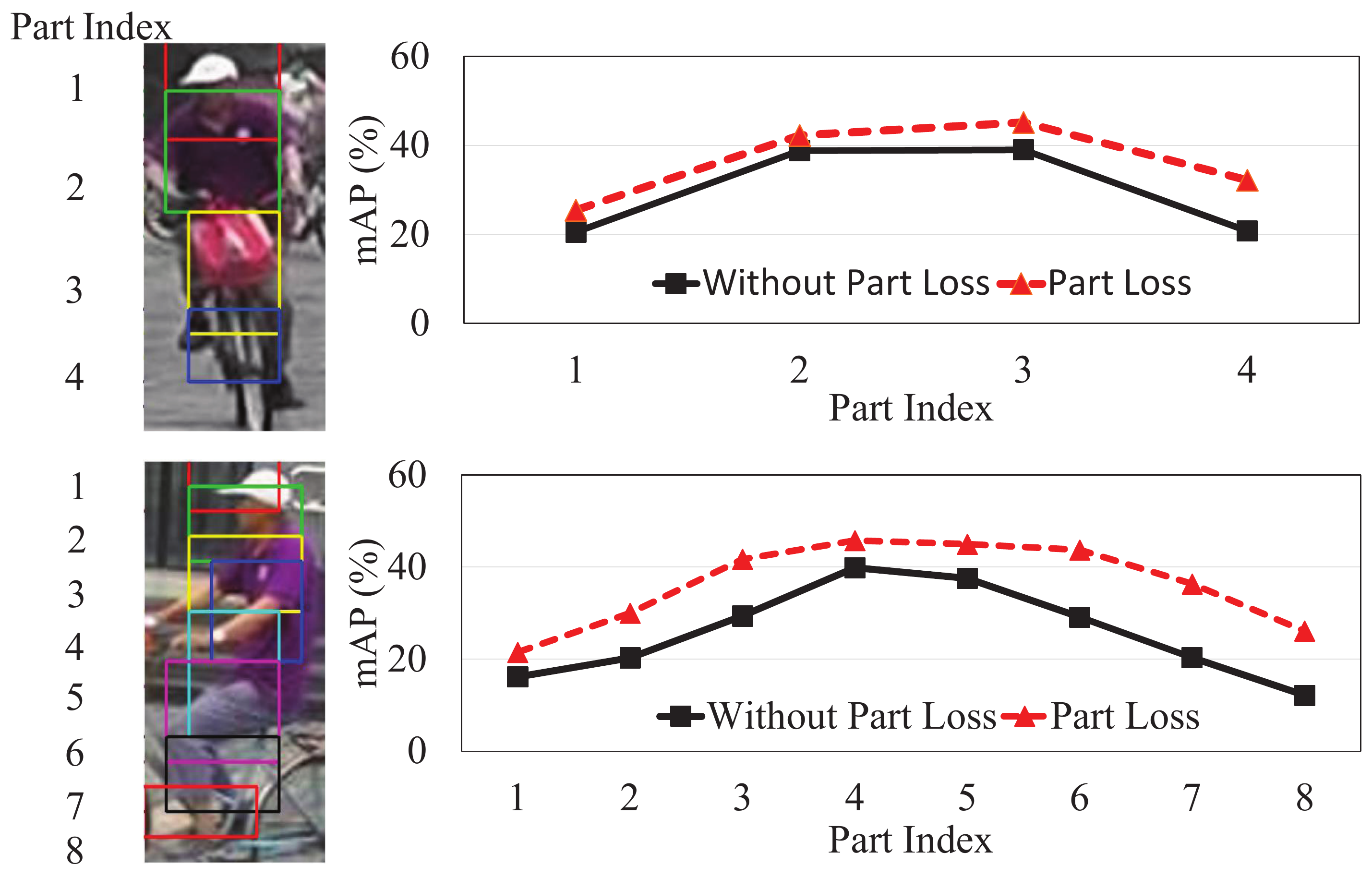}
\caption{Performance of part representation learned with and without part loss on Market1501.}
\label{Fig:generated_parts}
\end{figure}

\begin{table}
\centering
\footnotesize
\caption{Performance of final representations learned with our generated parts \emph{vs.} fixed grid parts with $K$=8 on Market1501.}
\begin{tabular}{lcc}
\hline
Part & mAP(\%) & Rank-1 (\%) \\
\hline
\hline
Grid Part&67.99 &86.96\\
Generated Part & 69.3&88.2\\
\hline
\end{tabular}
\label{Tab:comparison_grid}
\end{table}

\emph{Validity of Part Loss:} This experiment shows that part loss helps to minimize the representation learning risk and improve the descriptive power of CNN. We firstly show the effects of part loss computed with fixed grid parts. We equally divide an image into stripes, then learn part representations on them with and without part loss, respectively. We compare the ReID performance on Market1501. Fig.~\ref{Fig:local_description} clearly shows that more discriminative part representations can be learned with part loss for $K$ =4 and 8, respectively.

Besides using fixed grid part, we further perform experiments to show the validity of part loss computed on generated parts. Comparisons with similar settings are shown in Fig.~\ref{Fig:generated_parts}, where part loss also constantly improves the performance. Those two experiments show that, part loss enforces the network to learn more discriminative representations for different body parts, thus better avoids overfitting and decreases the representation learning risk for unseen person images.

\begin{table}
\centering
\footnotesize
\caption{Performance of global representation on Market1501 with different $K$. $K$=0 means the part loss is not considered.}
\begin{tabular}{ccccc}
\hline
$K$ &0& 2 & 4 &8 \\
\hline
\hline
mAP(\%) & 61.9&62.0&64.46&65.91\\
Rank-1 Acc.(\%) &81.5&81.9&84&85.6\\
\hline
\end{tabular}
\label{Tab:global_effect}
\end{table}
\emph{Performance of Global Representation:} This experiment verifies the effects of part loss to the global representation. As shown in Fig.~\ref{Fig:framework}, the global representation is computed on $\mathcal{X}$, which is also affected by the part loss. Experimental results on Market1501 are shown in Table~\ref{Tab:global_effect}, where $K$=0 means no part is generated, thus part loss is not considered. From Table~\ref{Tab:global_effect}, we could observe that part loss also boosts the global representation, \emph{e.g.,} the mAP and Rank-1 accuracy constantly increase with larger $K$. This phenomenon can be explained by the saliency maps in Fig.~\ref{Fig:illustration} (b), which shows the global representation learned with part loss focuses on larger body regions. We thus conclude that, part loss also boosts the global representation to focus on more body parts.

\begin{figure}
\begin{center}
\includegraphics[width=1\linewidth]{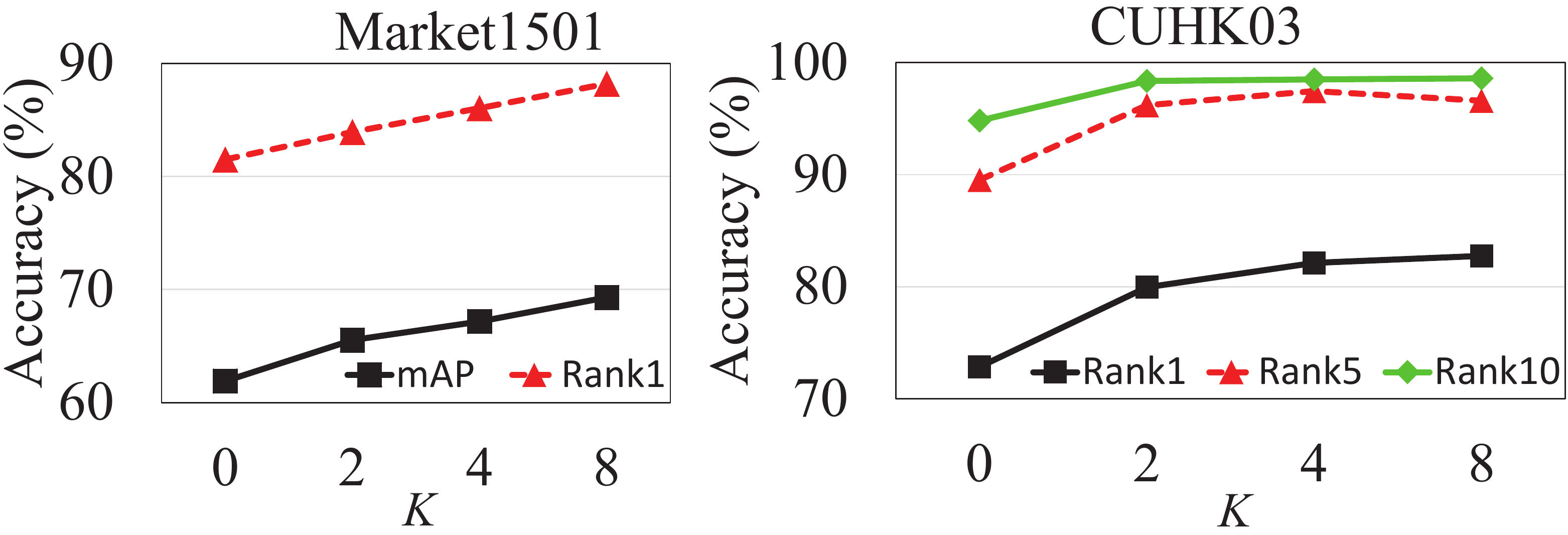}
\end{center}
\caption{Performance of final representation on Market1501 and CUHK03 with different $K$.}
\label{Fig:final_results}
\end{figure}

\emph{Performance of Final Representation:} $K$ is the only parameter for part loss. We thus test the performance of the final representation with different $K$. As shown in Fig.~\ref{Fig:final_results}, the final representation performs better with larger $K$, which extracts more detailed parts. This is consistent with the observation in Table~\ref{Tab:global_effect}. This also partially validates our part generation algorithm and part loss. Therefore, we set $K$=8 in the following experiments.

\emph{Discussions on Part Loss:} For Peron ReID, it is hard to directly model unseen person images because they are not given during training. We thus propose the part loss to decrease the representation learning risk on unseen person images. Part loss is a strict constraint, \emph{i.e.}, it is difficult to predict person ID from a single body part. By posting this strict constraints, we enforce the network to learn discriminative features for different parts, thus avoid overfitting on a specific part on the training set. As shown in the above experiments, the performance of a single part feature in Fig.~\ref{Fig:local_description} and Fig.~\ref{Fig:generated_parts} is not high. However, their concatenation achieves promising performance in Fig.~\ref{Fig:final_results}.

\begin{table}
\centering
\footnotesize
\caption{mAP achieved by different ways of part loss computation on Market1501. ``Concat." denotes part loss computed with concatenated part features. ``Final", ``Global", ``P-\emph{k}" denotes the final, global, and \emph{k}-th part representations. $K$ is set as 4.}
\begin{tabular}{ccccccc}
\hline
Methods & Final & Global & P-1 & P-2 & P-3 & P-4 \\
\hline
\hline
Concat. & 64.72&63.36& 21.80&38.55&37.78&19.39\\
\hline
Part Loss &67.17 & 64.46&25.43&42.24&45.19&32.19\\
\hline
\end{tabular}
\label{Tab:CompareConcat}
\vspace{-1.2em}
\end{table}

Our part loss is computed with Eq.~\eqref{Eq:partkloss},\emph{ i.e.}, compute the ID classification error on each part separately. Another possible solution is first to concatenate part representations then compute the ID classification with the fused features. We have compared those two methods and summarize the results in Table~\ref{Tab:CompareConcat}. As shown in the comparison, part loss computed with Eq.~\eqref{Eq:partkloss} performs better than the other solution, \emph{e.g.,} 67.17\%\emph{vs} 64.72\%. This might be because Eq.~\eqref{Eq:partkloss} better ensures the quality of each learned part feature, thus is more effective in decreasing the representation learning risk.

\subsection{Comparison with State-of-the-art}

In this section, we compare the proposed part loss networks (PL-Net) with existing ones on the Market1501, CUHK03, and VIPeR.

Table~\ref{Tab:comparison_market1501} shows the comparison on Market1501 in the terms of mAP and Rank-1 accuracy. As shown in Table~\ref{Tab:comparison_market1501}, the proposed method achieves the mAP of 69.3\% and Rank-1 accuracy 88.2\%, which both outperform existing methods. As shown in Table~\ref{Tab:comparison_market1501}, by adding the part loss, the global and part representation achieve 4\% and 7.1\% improvements in mAP over the baseline network, respectively. This makes the global and part representations already perform better than existing methods. By combining the global and part representations, PL-Net further boosts the performance.

\begin{table}
\centering
\footnotesize
\caption{Comparison on Market1501 with single query.}
\begin{tabular}{lcc}
\hline
Methods & mAP(\%) & Rank-1 (\%) \\
\hline
\hline
LOMO+XQDA~\cite{liao2015person} CVPR15 & 22.22 & 43.79\\
TMA~\cite{martinel2016temporal} ECCV16 & 22.31 & 47.92\\
DNS~\cite{zhang2016learning} CVPR16  & 35.68 & 61.02 \\
SSM~\cite{bai2017ssm} CVPR17 &68.80 & 82.21 \\
\hline
LSTM SCNN~\cite{varior2016siamese} ECCV16  & 35.31 & 61.60 \\
Gated SCNN~\cite{varior2016gated} ECCV16 & 39.55 & 65.88 \\
SpindleNet~\cite{zhao2017spindle} CVPR17  & - & 76.9 \\
MSCAN~\cite{li2017learning} CVPR17 &57.53 & 80.31\\
DLPAR~\cite{zhao2017deeply} ICCV17 & 63.4 & 81.0 \\
P2S~\cite{zhou2017point} CVPR17 & 44.27 & 70.72\\
CADL~\cite{lin2017consistent} CVPR17 &55.58 & 80.85\\
PDC~\cite{su2017pose} ICCV17 &63.41&84.14\\
\hline
Baseline Network &61.9&81.5\\
Global Representation &65.9&85.6\\
Part Representation &69&88.0\\
\textbf{PL-Net} & \textbf{69.3}&\textbf{88.2}\\
\hline
\end{tabular}
\label{Tab:comparison_market1501}
\vspace{-1.5em}
\end{table}

On CUHK03, the comparisons with existing methods are summarized in Table~\ref{Tab:comparison_cuhk03}. As shown in Table~\ref{Tab:comparison_cuhk03}, the global and part representations improve the baseline network by 8.1\% and  9.85\% on Rank-1 accuracy, respectively. The proposed PL-Net achieves 82.75\%, 96.59\%, and 98.59\% for the for Rank-1, Rank-5, and Rank-10 accuracies, respectively. This substantially outperforms most of the compared methods. Note that, the SpindelNet~\cite{zhao2017spindle} and PDC~\cite{su2017pose} are learned with extra human landmark annotations, thus leverages more detailed annotations than our method, and DLPAR~\cite{zhao2017deeply} has a higher baseline performance, \emph{e.g.,} 82.4\%~\cite{zhao2017deeply}  vs 72.85\% for our baseline.

The comparisons on VIPeR are summarized in Table~\ref{Tab:comparison_viper}. As VIPeR dataset contains fewer training images, it is hard to learn a robust deep representation. Therefore, deep learning-based methods~\cite{li2014deepreid,varior2016siamese,varior2016gated,qian2017multi,su2017pose} achieve lower performance than metric learning methods~\cite{bai2017ssm,chen2016similarity,matsukawa2016hierarchical,zhang2016learning}.  As shown in Table~\ref{Tab:comparison_viper}, simply using the generated representation obtains the Rank-1 accuracy of 47.47\%, which is lower than some metric learning methods~\cite{bai2017ssm,chen2016similarity,matsukawa2016hierarchical,zhang2016learning}. However, it outperforms most of recent deep learning based methods, \emph{e.g.}, DeepReID~\cite{li2014deepreid}, LSTM Siamese~\cite{varior2016siamese}, Gated Siamese~\cite{varior2016gated}, and MuDeep~\cite{qian2017multi}. Some recent deep learning based methods~\cite{zhao2017spindle,su2017pose,zhao2017deeply} perform better than ours. Note that, SpindelNet~\cite{zhao2017spindle} and PDC~\cite{su2017pose} leverage extra annotations during training. Also, the training set of DLPAR~\cite{zhao2017deeply} is larger than ours, \emph{i.e.,} the combination of \emph{CUHK03} and \emph{VIPeR}. Our learned representation is capable of combining with other features to further boost the performance. By combining the traditional LOMO~\cite{liao2015person} feature, we improve the Rank-1 accuracy to 56.65\%, which is the highest among all of the compared works.

From the above comparisons, we summarize : 1) part loss improves the baseline network and results in more discriminative global and part representations, and 2) the combined final representation is learned only with person ID annotations but outperforms most of existing works on the three datasets.

\begin{table}
\centering
\footnotesize
\caption{Comparison with existing methods on CUHK03.}
\begin{tabular}{lcccc}
\hline
Methods & Rank-1  & Rank-5  & Rank-10  \\
\hline
\hline
DeepReID~\cite{li2014deepreid} CVPR14 & 20.65 & 51.50& 66.5 \\
LSTM SCNN~\cite{varior2016siamese} ECCV16 & 57.3 & 80.1 & 88.3\\
Gated SCNN~\cite{varior2016gated} ECCV16  & 61.8 & 88.1 & 92.6 \\
DNS~\cite{zhang2016learning} CVPR16  & 62.55 & 90.05 & 94.80 \\
GOG~\cite{matsukawa2016hierarchical} CVPR16  & 67.3 & 91.0 & 96.0 \\
DGD~\cite{xiao2016learning} CVPR16  & 72.58 & 95.21 & 97.72 \\
SSM~\cite{bai2017ssm} CVPR17  & 76.6 & 94.6 & 98.0 \\
SpindleNet~\cite{zhao2017spindle} CVPR17  & 88.5 &97.8& 98.6 \\
MSCAN~\cite{li2017learning} CVPR17 & 74.21 & 94.33 & 97.54 \\
DLPAR~\cite{zhao2017deeply} ICCV17 & 85.4 & 97.6 & \textbf{99.4} \\
MuDeep~\cite{qian2017multi} ICCV17 & 76.87 & 96.12 & 98.41 \\
PDC~\cite{su2017pose} ICCV17  &\textbf{88.70}&\textbf{98.61}&99.24\\
\hline
Baseline Network &72.85&89.53& 94.82\\
Global Representation & 80.95&95.86&98.16\\
Local Representation &82.7&96.6& 98.59\\
\textbf{PL-Net} & 82.75&96.59&98.6\\
\hline
\end{tabular}
\label{Tab:comparison_cuhk03}
\end{table}

\begin{table}
\centering
\footnotesize
\caption{Comparison with existing methods on VIPeR.}
\begin{tabular}{lcccc}
\hline
Methods & Rank-1 & Rank-5  & Rank-10   \\
\hline
\hline
DNS~\cite{zhang2016learning} CVPR16 & 41.01 & 69.81 & 81.61 \\
TMA~\cite{martinel2016temporal} ECCV16 & 48.19 & 87.65 & 93.54 \\
GOG~\cite{matsukawa2016hierarchical} CVPR16  &49.72& 88.67 & 94.53\\
Null~\cite{zhang2016learning} CVPR16 & 51.17 & 90.51 & 95.92 \\
SCSP~\cite{chen2016similarity} CVPR16  & 53.54 & \textbf{91.49} & \textbf{96.65} \\
SSM~\cite{bai2017ssm} CVPR17  & 53.73 & 91.49 & 96.08 \\
\hline
DeepReID~\cite{li2014deepreid} CVPR14 & 19.9 & 49.3 & 64.7 \\
Gated Siamese~\cite{varior2016gated} ECCV16 & 37.8 & 66.9 & 77.4 \\
LSTM Siamese~\cite{varior2016siamese} ECCV16  & 42.4 & 68.7 & 79.4 \\
SpindleNet~\cite{zhao2017spindle} CVPR17 &53.8 & 74.1 & 83.2 \\
MuDeep~\cite{qian2017multi} ICCV17 & 43.03 & 74.36 & 85.76 \\
DLPAR~\cite{zhao2017deeply} ICCV17  & 48.7 & 74.7 & 85.1 \\
PDC~\cite{su2017pose} ICCV17 &51.27&74.05&84.18\\
\hline
Baseline Network&34.81&61.71&72.47\\
Global Representation & 44.30 & 69.30 & 79.11 \\
Local Representation & 44.94& 72.47&80.70\\
PL-Net & 47.47&72.47&80.70 \\
\textbf{PL-Net+LOMO~\cite{liao2015person}} & \textbf{56.65}&82.59&89.87 \\
\hline
\end{tabular}
\label{Tab:comparison_viper}
\vspace{-1.5em}
\end{table}

\section{Conclusions}
This paper shows that, the traditional deep classification models are trained with empirical classification risk on the training set. This makes those deep models not optimal for representation learning in person ReID, which can be regarded as a zero-shot learning problem. We thus propose to minimize the representation learning risk to infer more discriminative representations for unseen person images. The person part loss is computed to evaluate the representation learning risk. Person part loss firstly generates $K$ body parts in an unsupervised way, then optimizes the classification loss for each part separately. In this way, part loss network learns discriminative representations for different parts. Extensive experimental results on three public datasets demonstrate the advantages of our method. This work explicitly infers parts based on the given parameter $K$. More implicit ways to minimize the representation learning risk will be explored in our future work.
{\small
\bibliographystyle{ieee}
\bibliography{egbib}
}
\end{document}